\documentclass[letterpaper, 10 pt, conference]{ieeeconf} 
\pdfoutput=1   
\IEEEoverridecommandlockouts
\usepackage[backend=biber,
            hyperref=true,
            url=false,
            isbn=false,
            doi=false,
            backref=false,
            style=ieee,
            natbib=true,
            mincitenames=1,
            maxcitenames=1,
            citestyle=numeric-comp,
            sorting=nyt,
            block=none]{biblatex}
                
\renewcommand{\bibfont}{\small}
\usepackage{flushend}
\usepackage{balance}
\usepackage{graphics}
\usepackage[pdftex]{graphicx}
\usepackage{wrapfig}
\DeclareGraphicsExtensions{.pdf,.png,.jpg}
\usepackage{epsfig}
\usepackage[font={small}]{caption}
\usepackage{subcaption}
\usepackage[rightcaption]{sidecap}
\usepackage{pbox}

\usepackage{bigstrut}
\usepackage{mathtools}
\usepackage{amsmath, amssymb, amscd}
\usepackage{ wasysym } 
\usepackage{amsfonts}
\usepackage{mathptmx} 
\usepackage{gensymb} 
\usepackage{nicefrac}       
\numberwithin{equation}{section} 

\DeclareMathAlphabet{\mathcal}{OMS}{lmsy}{m}{n}
\DeclareSymbolFont{largesymbols}{OMX}{cmex}{m}{n}
\usepackage{textcomp} 

\usepackage{algorithm} 
\usepackage{algorithmicx, algpseudocode} 

\usepackage{array} 
\usepackage{tabularx}
\usepackage{multirow}
\usepackage{multicol}
\usepackage{booktabs}
\usepackage{tabulary}
\usepackage{colortbl}

\usepackage[T1]{fontenc} 
\usepackage[utf8]{inputenc}
\usepackage[english]{babel} 
\usepackage{units}
\usepackage{bm}
\usepackage{times} 
\usepackage{xspace}
\usepackage{flushend}
\usepackage{balance} 
\usepackage{csquotes}
\usepackage{makeidx}
\usepackage{blindtext}

\let\labelindent\relax
\usepackage{enumitem}

\usepackage{tikz}
\usetikzlibrary{backgrounds}

\usepackage{ragged2e}
\usepackage{soul} 
\usepackage{subfiles} 

\usepackage[protrusion=true,expansion=true]{microtype}
\usepackage[yyyymmdd]{datetime}

\date{\protect\formatdate{1}{1}{2001}}

\usepackage{url}
\makeatletter
\g@addto@macro{\UrlBreaks}{\UrlOrds}
\makeatother
\usepackage{pgfplots}
\pgfplotsset{compat=newest}
\usepackage{color}

\usepackage[bookmarks=false]{hyperref}
\hypersetup{
    colorlinks=true,
    linkcolor=black,
    citecolor=black,
    filecolor=cyan,
    urlcolor=black
}

\usepackage{marginnote}
\usepackage{soul} 

\usepackage[colorinlistoftodos,prependcaption,textsize=small]{todonotes}
\usepackage{regexpatch}
\makeatletter
\xpatchcmd{\@todo}{\setkeys{todonotes}{#1}}{\setkeys{todonotes}{inline,#1}}{}{}
\makeatother

\newcommand{\tocite}[1]{%
\textcolor{red}{[cite:\ifthenelse{\equal{#1}{}}{}{#1}?]}
}

\newcommand{\ignore}[1]{}

\renewcommand{\baselinestretch}{1.0}



\newcommand{\anig}[1]{\todo[inline,color=blue!20]{AG: #1}}

\title{\LARGE \bf
Scaling Robot Supervision to Hundreds of Hours with RoboTurk: \\Robotic Manipulation Dataset through Human Reasoning and Dexterity
}

\author{Ajay Mandlekar$^{1}$, Jonathan Booher$^{1}$, Max Spero$^{1}$, Albert Tung$^{1}$, Anchit Gupta$^{1}$, \\Yuke Zhu$^{1}$, Animesh Garg$^{2,3}$, Silvio Savarese$^{1}$, Li Fei-Fei$^{1}$
\thanks{$^{1}$Stanford AI Lab (SAIL), Stanford University. $^{2}$Nvidia, USA. $^{3}$University of Toronto, Canada.
Email: {\tt\small amandlek@stanford.edu}}%
}

\begin{document}

\maketitle
\thispagestyle{empty}
\pagestyle{empty}

\begin{abstract}
Large, richly annotated datasets have accelerated progress in fields such as computer vision and natural language processing, but replicating these successes in robotics has been challenging. While prior data collection methodologies such as self-supervision have resulted in large datasets, the data can have poor signal-to-noise ratio. By contrast, previous efforts to collect task demonstrations with humans provide better quality data, but they cannot reach the same data magnitude. Furthermore, neither approach places guarantees on the diversity of the data collected, in terms of solution strategies. In this work, we leverage and extend the RoboTurk platform to scale up data collection for robotic manipulation using remote teleoperation. The primary motivation for our platform is two-fold: (1) to address the shortcomings of prior work and increase the total quantity of manipulation data collected through human supervision by an order of magnitude without sacrificing the quality of the data and (2) to collect data on challenging manipulation tasks across several operators and observe a diverse set of emergent behaviors and solutions. We collected over 111 hours of robot manipulation data across 54 users and 3 challenging manipulation tasks in 1 week, resulting in the largest robot dataset collected via remote teleoperation. We evaluate the quality of our platform, the diversity of demonstrations in our dataset, and the utility of our dataset via quantitative and qualitative analysis. For additional results, supplementary videos, and to download our dataset, visit \href{http://roboturk.stanford.edu/realrobotdataset}{\texttt{roboturk.stanford.edu/realrobotdataset}}
\end{abstract}

\section{INTRODUCTION}










Crowdsourcing mechanisms such as Amazon Mechanical Turk have facilitated the creation of large, richly annotated datasets. The advent of datasets, sizing in millions, has accelerated progress in computer vision and natural language processing~\cite{deng2009imagenet,rajpurkar2018squad2} by enabling the development and evaluation of a wide range of learning algorithms and benchmarks. Efforts to aggregate similar amounts of data promise to boost performance in the field of robot manipulation.

Subsequently, the community leveraged online self-supervised data collection ~\cite{levine2016learning, pinto2016supersizing} and off-policy reinforcement learning ~\cite{kalashnikov2018qt} to collect large quantities of physical robot data for tasks such as grasping (over 1000 hours).
However, the data collected through such methods often has low signal-to-noise ratio, since a large portion of the data is collected by applying random controls. 
Subsequently, the time it takes to start collecting high quality data can be prohibitively large, limiting the complexity of the tasks achievable with this approach.
Furthermore, specification and evaluation of a reward function for complex tasks can be non-intuitive. In contrast, human demonstrations obviate the need for this specification by implicitly providing a set of successful task executions.  

\begin{figure}[!t]
    \centering
    \includegraphics[width=\linewidth]{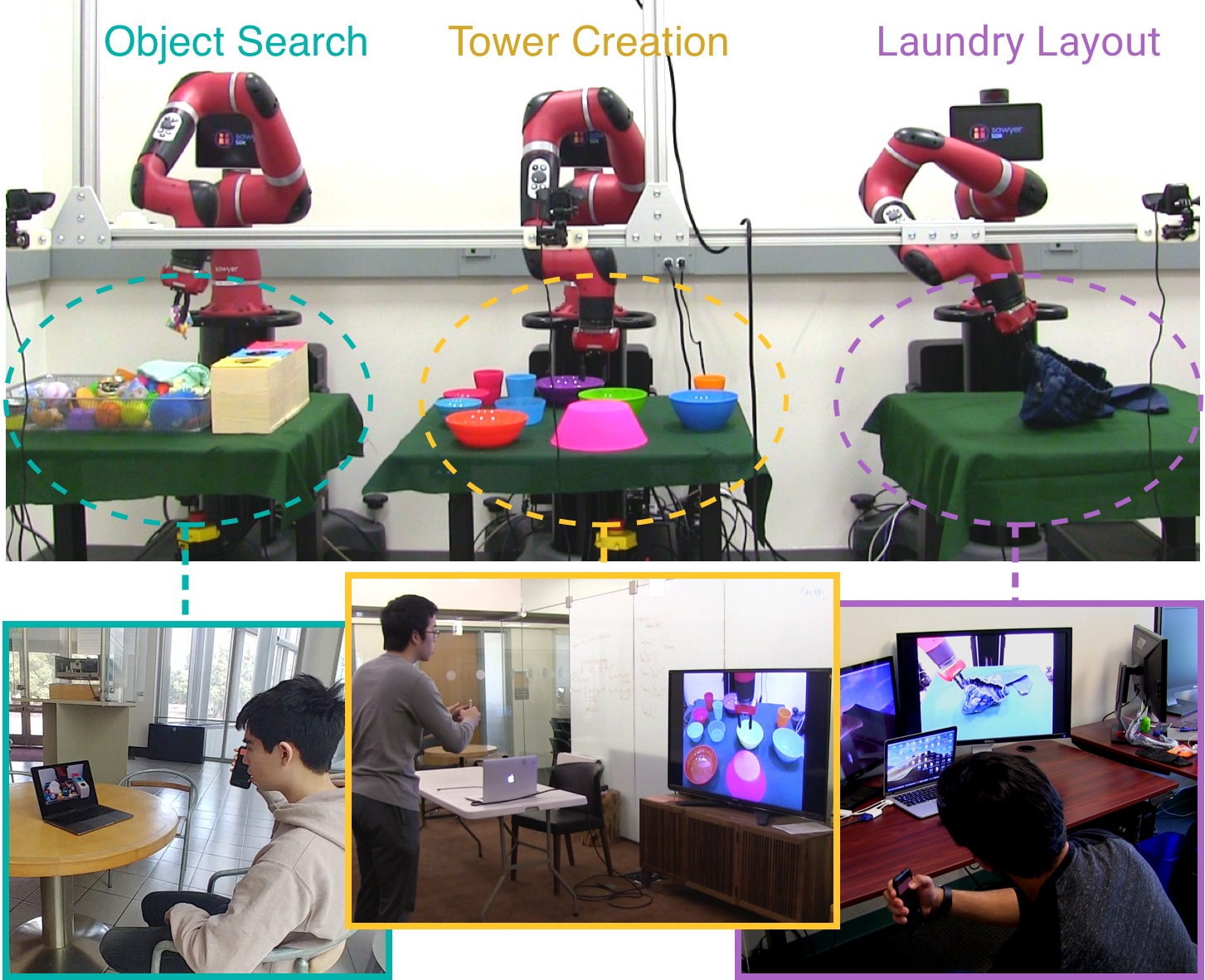}
    \caption{\textbf{Collecting data on physical robot arms with the RoboTurk platform.} To collect task demonstrations, users connect to our platform from remote locations using a web browser and use their smartphone as a motion controller to move the physical robot arm in free space. Users are provided a video stream of the robot workspace in their web browser.}
    \label{fig:intro}
    \vspace{0.5mm}
\end{figure}

Prior work ~\cite{zhang2017deep,forbes2014robot} has shown that imitation learning on data collected from humans can achieve success on a set of restricted task instances~\cite{argall2009survey,pomerleau1989alvinn,abbeel2010autonomous,boularias2011relative,krishnan2019swirl,ross2013learning,vevcerik2017leveraging}.
However, these approaches have been limited in both the scale of data collected and the complexity of the tasks used for data collection. The ideal approach would be able to collect data on the scale of self-supervised methods but with the quality of human-in-the-loop approaches. 

However, replicating the success and impact that large-scale datasets have had in vision and language for robotics has been challenging. The core problem is that the expert needs to \textit{demonstrate} how to perform a task in real-time, instead of offline data-labeling. Therefore, methods for real-time remote interaction that are robust to delays in both actuation and network latency must be established. More importantly these methods must operate at scale to facilitate crowdsourcing. 

This paper addresses the problem of large scale crowdsourcing on real robots. We propose an approach to collect task demonstrations on physical robots from humans to scale data collection along two dimensions: the quantity of data collected, and the diversity of data collected. We extend the RoboTurk platform~\cite{mandlekar2018roboturk} from simulation to real robots, and address the ensuing challenges such as: establishing a framework to remotely manage robots, handling additional delays due to hardware actuation not present in simulators, ensuring operational safety when operated by novice users, and managing alignment of data streams from a multitude of sensors for a number of robots simultaneously.
Furthermore, we present three robotic manipulation tasks that require human intervention both at the level of reasoning and dexterity of manipulation. Lastly, we present a large-scale dataset on these three tasks, comprised of over 111 hours of data collected by 54 people resulting in a diverse set of solutions that is an order of magnitude larger than state-of-the-art. 

\vspace{1mm}
\noindent \textbf{Summary of Contributions:}
\begin{enumerate}[
    topsep=0pt,
    noitemsep,
    partopsep=0.5ex,
    parsep=0.5ex,
    leftmargin=*,
    itemindent=3ex
    ]
\item We extend the RoboTurk crowdsourcing platform to enable remote data collection on physical robots. Our new platform accounts for limited robot resources, additional delays introduced by robot hardware, the need for safety measures to protect the robots from harm, and data collection from multiple sensor streams at different rates. 
\item We introduce Object Search, Tower Creation, and Laundry Layout: three different tasks that require human intervention in the form of low-level dexterity of manipulation and high-level cognitive reasoning to solve the tasks.
\item We present the largest robot dataset collected via remote teloperation. Over the course of 1 week, we collected over \textit{111 hours} of data across \textit{54 users} on the 3 challenging manipulation tasks that we introduced.
\item We evaluate the quality of our platform, the diversity of demonstrations in our dataset, and the utility of our dataset via quantitative and qualitative analysis.
\end{enumerate}

\section{RELATED WORK}

\noindent \textbf{Large-Scale Data collection in Robotics.}
Data-driven methods for learning in robotics have been used to collect grasps~\cite{goldfeder2009columbia} and object models~\cite{kasper2012kit}, and run large scale physical trials for grasping~\cite{levine2016learning,pinto2016supersizing,kalashnikov2018qt} and pushing~\cite{yu2016more}. These methods used hundreds of hours of robot interaction, although a majority of the trials were not successful.

\noindent \textbf{Simulated and Self-supervised Methods.} 
Large scale self-supervision has low signal-to-noise ratio due to exploration via a random policy. 
While simulators can scale easily and provide many task variations, several task types, such as those shown in this work, can be difficult to simulate. Combinations of these methods as in~\cite{mahler2017dex, james2019sim}, are limited by simulator fidelity, and often focused on tasks with specific and easily measurable success criterion. 

\noindent \textbf{Learning from Demonstration and Imitation Learning.} 
Imitation learning (IL) is often preferred over RL to achieve efficiency in policy learning.
Specification of reward functions can be non-intuitive for a number of robotic tasks~\cite{ng2000algorithms}.
Imitation learning can be performed mainly through inverse reinforcement learning (IRL)~\cite{abbeel2011inverse,krishnan2019swirl} or behavioral cloning (BC)~\cite{pomerleau1989alvinn,ross2013learning, schulman2016learning}. 
However, these algorithms typically either require a large amount of data (BC) or a large number of environment interactions (IRL). 


\begin{figure}[t!]
    \centering
    \vspace{1mm}
    \includegraphics[width=\linewidth]{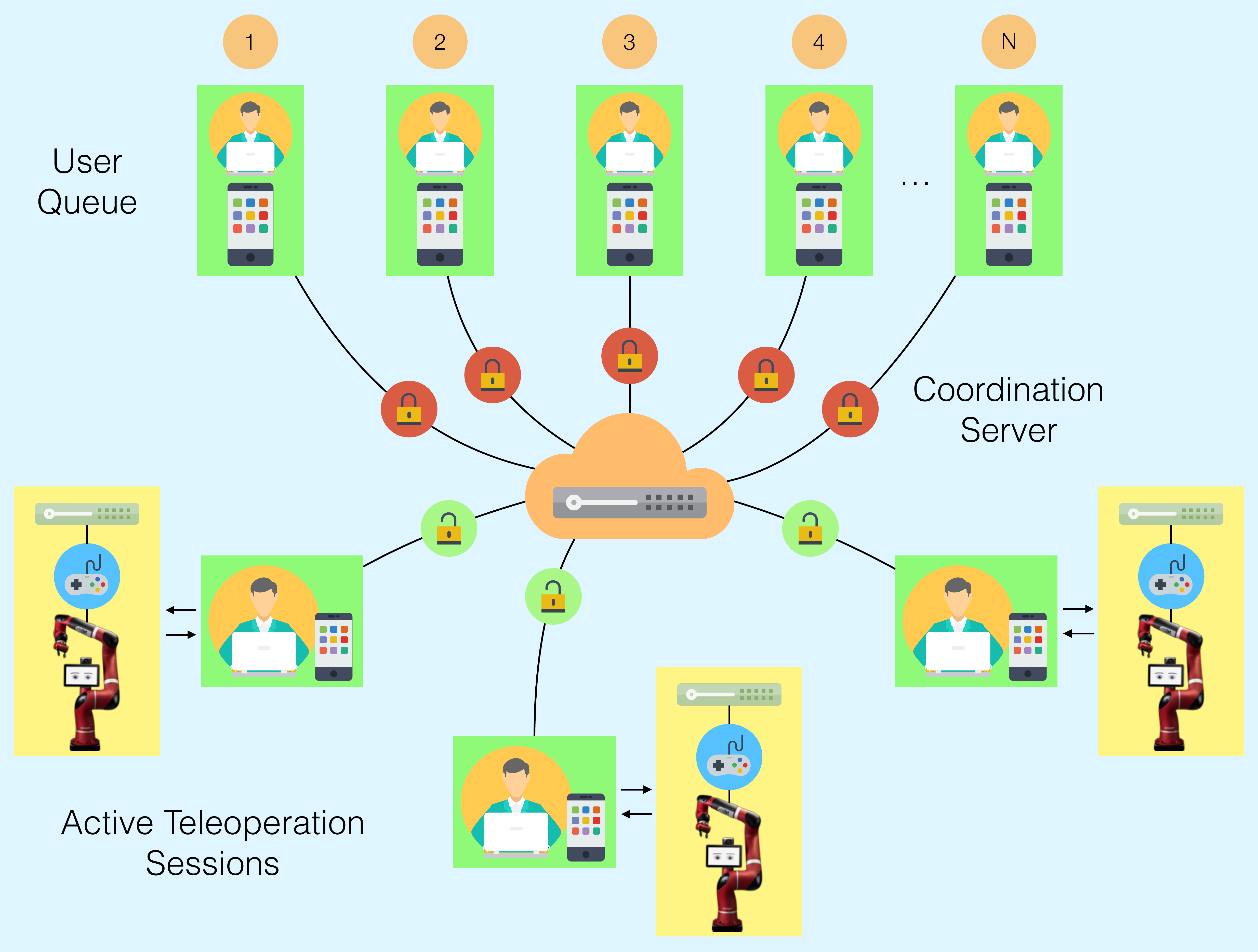}
    \caption{\textbf{RoboTurk system diagram for data collection on physical robot arms.} The coordination server manages users and guards access to the physical robot arms. There is a limited number of active teleoperation sessions (one per robot arm). Other users are queued in the system until one of the robot arms is available for data collection. When this occurs, the coordination server creates a new connection between the robot arm and the user at the front of the queue, and a new teleoperation session begins.}
    \label{fig:system-diagram}
\end{figure}

\begin{figure*}[t!]
    \centering
    \includegraphics[width=0.9\linewidth]{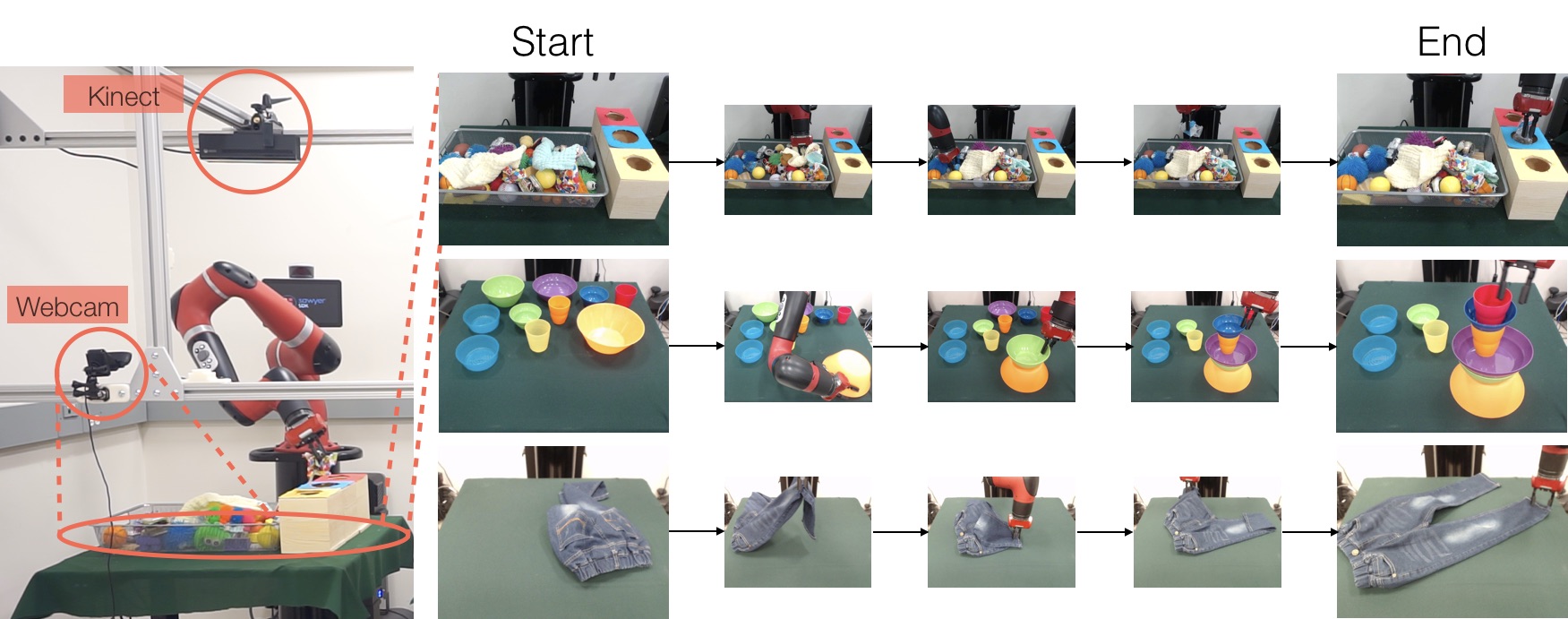}
    \caption{\textbf{Data Collection Setup and Tasks:} Our data collection setup (left) consisted of three Sawyer robot arms, each of which had a front-facing webcam view and a top-down Kinect depth camera view. The front-facing view was streamed to each teleoperator. We collected data on three tasks (right) that require both fine-grained dexterity and high-level planning to solve. In the \textit{Object Search} task (top) the objective is to find all instances of a certain target object category (plush animal, plastic water bottle, or paper napkin) and fit them into the corresponding bin. In the \textit{Tower Creation} task (middle), the objective is to stack the various cups and bowls to create the tallest tower. In the \textit{Laundry Layout} task (bottom) the objective is to layout an article of clothing on the table such that it lies flat without folds.}
    \label{fig:tasks}
\end{figure*}

\noindent \textbf{Crowdsourced Teleoperation for Robot Learning.}
Collecting large amounts of data has been a challenge for continuous manipulation tasks. Crowdsourcing supervision has resulted in some remarkable scaling of datasets in computer vision and natural language \cite{deng2009imagenet,rajpurkar2018squad2}.  
In robotics, crowdsourcing over the internet was first introduced to robotics in the Telegarden Project~\cite{Goldberg94beyondthe}.
Since then a number of studies have leveraged the crowd to \textit{ask for help}~\cite{Sorokin-2010-10559,sung2018robobarista,hokayem2006bilateral}. Prior works have also built frameworks for web-based low-latency robot control~\cite{toris2014robot}.  Kehoe et al.\cite{kehoe2014tase} provides a treatise that touches on the aspects of cloud robotics: big data, cloud computing, collective robot learning and crowdsourcing. 
Teleoperation mechanisms vary from game interfaces~\cite{marin2005multimodal} to free-space positioning interfaces~\cite{kofman2005teleop}. A comparison of various control interfaces shows that general purpose hardware is deficient while special purpose hardware is more accurate but is not widely available~\cite{kent2017comparison, marin2005multimodal}. 

Virtual reality-based free-space controllers have recently been proposed both for data collection~\cite{whitney2017comparing,lipton2018baxter} and policy learning~\cite{yan2017learning,zhang2017deep}. 
While these methods have shown the utility of data, they do not provide a seamlessly scalable data collection mechanism. 
Often the data is either collected locally or requires a powerful local client computer, to render the high definition sensor stream to a VR headset~\cite{whitney2017comparing, zhang2017deep}.
The use of VR hardware and requirement of client-side compute resources has limited the deployment of these interfaces on crowdsourcing platforms. 

Our system builds on RoboTurk~\cite{mandlekar2018roboturk}, which uses a ubiquitous smartphone-based 6-DoF controller along with seamless cloud integration to ensure homogeneous quality of service regardless of client's compute resources.
In contrast to local teleoperation methods that restrict data collection to a few users, crowdsourcing mechanisms such as RoboTurk can allow several interesting strategies to be demonstrated that vary across people, and across situations, leading to \textit{diversity} of the data collected, as shown in Fig.~\ref{fig:multimodal}.

\section{SYSTEM DESIGN}

In order to collect our dataset, we leveraged the RoboTurk platform \cite{mandlekar2018roboturk}, which allows large groups of remote users to simultaneously collect task demonstrations by using their smartphones as motion controllers to control robots in simulated domains. We first review the original platform and then discuss the extensions we implemented to enable robust data collection on physical robot arms.

\subsection{RoboTurk Overview} 
RoboTurk is a platform that allows users to seamlessly collect task demonstrations in simulation through low-latency teleoperation, regardless of their location or compute resources. Users connect to a website that streams video from the simulated environment, and use their smartphone as a motion controller to control the robot. The simulation itself runs in a remote server hosted in the cloud -- this is to ensure homogeneous quality of service to every user regardless of available compute resources. In this way, RoboTurk facilitates crowdsourcing task demonstrations in simulated domains from large pools of remotely located annotators. 


In order to support concurrent low-latency teleoperation servers that allow many users to use the platform simultaneously, the platform utilizes several core components. RoboTurk leverages Web Real-Time Communication (WebRTC) to establish low-latency communication links between a user's phone, web browser, and a teleoperation server that is hosted in the cloud. We now outline the core components.

\vspace{1mm}
\noindent \textbf{User Endpoint.} 
Each user controls the robot arm using their smartphone as a motion controller and receives a real-time video stream of the robot workspace in their web browser. The phone transmits its changes in position and its current absolute orientation over WebRTC to the teleoperation server. 
We additionally provide users with the ability to enable and disable control of the robot so that users can re-position their arm and phone for better ergonomics similar in nature to interfaces for surgical robots \cite{fritsche2015first}.

\noindent \textbf{Coordination Server.} This server creates and manages teleoperation sessions when users enter or leave the system. 

\noindent \textbf{Teleoperation Server.} This is a process dedicated to a single user that is created by the centralized coordination server on a per-user basis. The server maintains its own simulator instance, sends frames to the user's web browser, and handles the incoming control commands from the user.

\begin{table*}[t!]
    \centering
    \vspace{4mm}
    \caption{\textbf{Dataset Comparison.} We compare our dataset to similar robot datasets collected via human supervision in prior work. Items marked with $*$ are estimates that were extrapolated using other reported information, and interfaces marked with $^{\dagger}$ are not real-time.}
    \resizebox{0.9\linewidth}{!}{%
 \begin{tabular}{c c c c c c} 
 \hline
 \rowcolor[HTML]{CBCEFB}
 Name & Interface & Task& Avg. Task Length (sec)& Number of Demos & Total Time (hours) \\ [0.2ex] 
 \hline
JIGSAWS\cite{JIGSAWS}& daVinci & surgery & 60* & 103 &  1.66\\
\rowcolor[HTML]{EFEFEF} 
Deep Imitation \cite{zhang2017deep} & VR &pick, grasp, align & 5* & 1664 & 2.35 \\
DAML\cite{DAML} & Human demos & pick, place, push & 5* & 2941& 4.08 \\
\rowcolor[HTML]{EFEFEF}
MIME\cite{MIME} & Kinesthetic&pick, place, push, pour& 6* & 8260 & 13.7* \\
PbD\cite{forbes2014robot} & GUI$^{\dagger}$ & pick, place& 207*& 465 & 25.8* \\
\rowcolor[HTML]{EFEFEF} 
Roboturk-Real (Our) & iPhone AR & long horizon object manip & 186 & 2144 & \textbf{111.25}\\ \hline
\end{tabular}
}
\label{fig:dataset_comparison}
\end{table*}

\subsection{Extending RoboTurk for use with real robots} While the initial RoboTurk platform works well for collection in simulated domains, collecting data on physical robots poses additional challenges. 

\vspace{1mm}
\noindent \textbf{Limited resources.} 
While the number of simulation instances that can be run in parallel is bottlenecked only by compute resources, data collection on physical robots is bottlenecked by the number of available robots. 
Thus, to extend RoboTurk to the physical robot setting, we implemented a mutual exclusion principle to limit the number of users on the platform. The coordination server loads a centralized landing page for all users and routes them to a robot and corresponding task. It places a lock on the control of each physical robot arm so that only one user may operate the robot at a time. 
    
\noindent \textbf{Control latency.} Controlling a robot in a simulator is markedly different from controlling a physical robot due to stochastic delays that can occur at a hardware level (e.g. commanding the motors) and at a network level, since commands are sent to the robot through a network connection. These are uncontrollable delays that are incurred in addition to those from the connection between the user and the teleoperation server. To account for these delays, we use a low-pass filter to reject high-frequency user input to ensure smooth and responsive teleoperation.
By focusing on the low-frequency content in a user's command stream, we are able to ensure that delays do not adversely affect a person's capability to control the robot arm. Furthermore, instead of having teleoperation servers run on cloud infrastructure, we now spawn teleoperation servers on machines that are located in close physical proximity to the robots in order to minimize the latency of control commands send to the robot arms.
    
\noindent \textbf{Robot safety.} 
We address the need for safety of the robot arms and workspaces by extending the RoboTurk smartphone app to ensure that the user is holding the phone correctly and exercising slow, deliberate motions by validating phone poses.
Participants in our data collection study were given a structured 5 minute tutorial to familiarize themselves with operation of the physical robot arm, and also given a direct line of communication to people monitoring the robots to ensure quick responses to unsafe user control. 
    
\noindent \textbf{Data collection.} Data collection in simulation is straightforward, since the state of the simulator can be saved at each timestep. From this minimal simulator state, all ground-truth observations of interest can be reconstructed. In contrast, data collection in the real world is much more unstructured. Multiple sensor streams emit data at different rates. To account for this, we leverage rosbag, a package built on top of the Robot Operating System (ROS), that allows for recording messages sent over concurrent publisher-subscriber communication channels called topics.

\section{DATA COLLECTION}

\subsection{Task Design}

We designed three robotic manipulation tasks for data collection. These tasks were chosen with care - each task requires the use of low-level dexterity and high-level reasoning in order to solve it - both of which can be provided with a human in the loop. Furthermore, the solution space is multimodal - there are several admissible ways to solve the task for a given task instance. Consequently, there is inherent freedom in our tasks, encouraging our diverse set of users to experiment with different solution strategies.

\vspace{1mm}
\noindent \textbf{Object Search.} The goal of this task is to search for a set of target objects within a cluttered bin and fit them into a specific box. There are three target object categories: \textit{plush animals}, \textit{plastic water bottles}, and \textit{paper napkins}. The workspace consists of a large cluttered bin containing a diverse assortment of objects and three target boxes, one per category of target object. At the start of each task instance, three target objects of each category are mixed in among the clutter of the box. A target category is randomly sampled and relayed to the operator, who must use the robot arm to find all three objects corresponding to the target category and place each item into its corresponding hole. We further place the constraint that objects can be grasped and moved around within the bin but they cannot be placed outside the bin in any sort of staging area - this adds to the challenging nature of the task. 


The Object Search task requires human-level reasoning to detect and search for the target items and dexterous manipulation to dig through the bin, push objects out of the way, pick up the target object successfully, and fit the target object into the corresponding hole, making it a good candidate for crowdsourcing. The objects also have interesting properties - the paper napkins appear in crumpled and unfolded configurations, and the crushed plastic water bottles are challenging to detect and grasp due to their translucence and arbitrary rigid shape. Furthermore, it is a practical problem with industrial applications~\cite{danielczuk2019mechsearch}.





\noindent \textbf{Tower Creation.} In this task, an assortment of cups and bowls are arranged on the table. The goal of the task is to create the tallest tower possible by stacking the cups and bowls on top of each other. This task requires physical reasoning over the properties of each type of cup and bowl and thinking about how to stack them on top of each other to maximize height without sacrificing the stability of the tower. 



We diversify the initial task configurations by sampling a set of ten objects drawn from a total of 28 bowls in 7 varieties and 12 cups in 3 varieties. We also randomize the initial configuration of the objects. This encourages diversity in the demonstrations since users will not receive the same set of objects in the same configuration, enforcing each demonstration to be unique. 


\begin{figure*}[t!]
    \includegraphics[width=\linewidth]{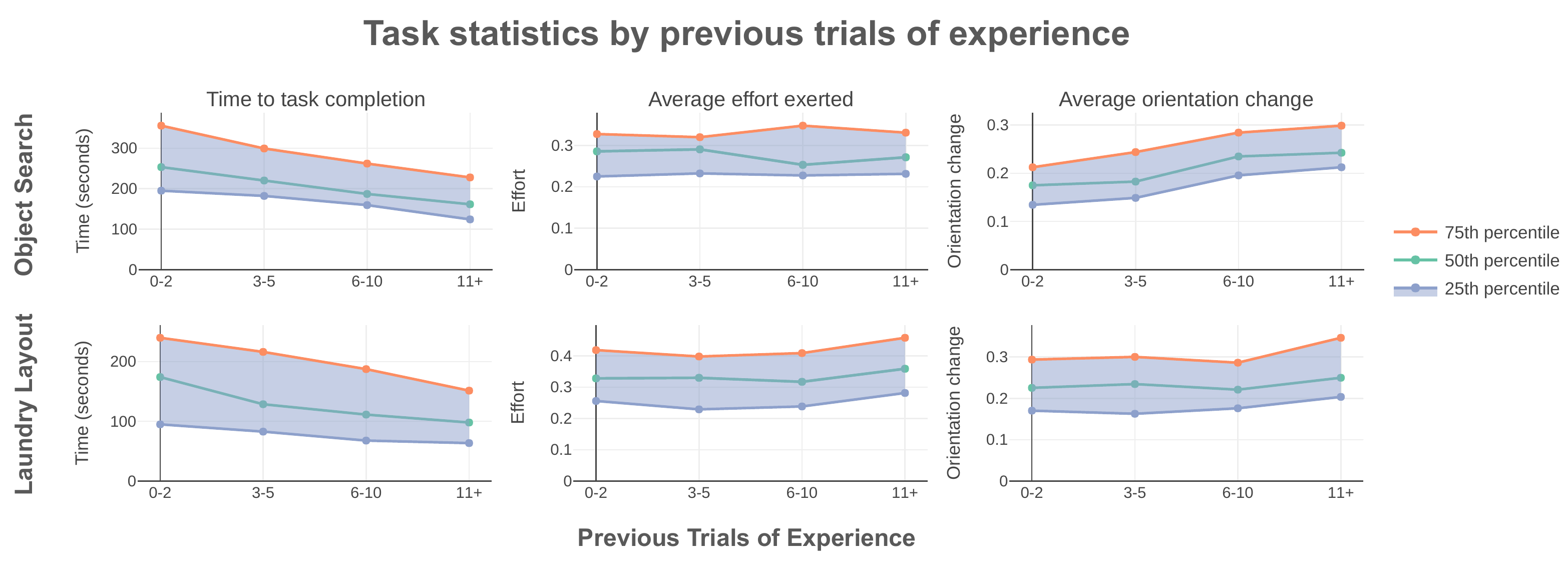}
    \caption{\textbf{Characterizing user skill improvement over time.}
    Task completion times across users versus number of demonstrations of experience (left), average effort exerted versus experience (middle), and average change in orientation versus experience (right) on Object Search (top) and Laundry Layout (bottom). Together, these show that with experience, users learn to use more nuanced motion to complete tasks faster and more efficiently. Task completion time drops steadily as experience increases. However, average effort exerted (measured by the square $L_2$ norms of phone translations) is largely invariant across experience, while phone orientation change increases with experience, implying that users improve over time not by moving the phone faster, but rather by learning to enact more dexterous motions. Each graph displays quartiles to show that changes are consistent across the entire population. The Tower Creation task was excluded from this evaluation since most users insisted on using all five minutes to create their tower.}
    \label{fig:experience_graphs}
\end{figure*}

\noindent \textbf{Laundry Layout.} This task starts with a hand towel, a pair of jeans, or a t-shirt placed on the table. The goal is to use the robot arm to straighten the item so that it lies flat on the table with no folds. On every task reset we randomly place the item into a new configuration. This task was chosen for the visual and physical reasoning skills necessary to unfold and flatten the item. Solving it requires understanding the current item configuration and how it will respond to different types of contact. 



\subsection{Data Collection and Dataset Details}

We collected our dataset using 54 different participants over the course of 1 week. Every user participated in a supervised hour of remote data collection, including a brief 5 minute tutorial at the beginning of the session. Afterwards, they were given the option to collect data without supervision for all subsequent collection. The users who participated in our data collection study collected the data from a variety of locations. All locations were remote - no data collection occurred in front of the actual robot arms. 

Fig.~\ref{fig:tasks} shows our data collection setup. We collected data on three Sawyer robot arms - each of which had a front-facing webcam and a top-down Kinect depth camera mounted in the workspace of the robot arm. We collected RGB images from the front-facing RGB camera (which is also the teleoperator video stream view) at 30Hz, RGB and Depth images from the top-down Kinectv2 sensor also at 30Hz, and robot sensor readings such as joint and end effector information, at approximately 100Hz.

Table~\ref{fig:dataset_comparison} compares our dataset against other robot datasets collected using humans. With over 111 hours of total robot manipulation data in our dataset, our dataset is 1-2 orders of magnitude larger than most other datasets. The number of task demonstrations in our dataset also compares favorably with the number of demonstrations in large datasets such as MIME~\cite{MIME}, but the tasks that we collected data on are more difficult to complete, as they take on the order of minutes to complete successfully, as opposed to seconds (see Fig.~\ref{fig:time_per_user}).

\section{EXPERIMENTS: SYSTEM ANALYSIS}

\subsection{Quantitative User Performance Analysis}

\begin{figure}[h!]
    \includegraphics[width=\linewidth]{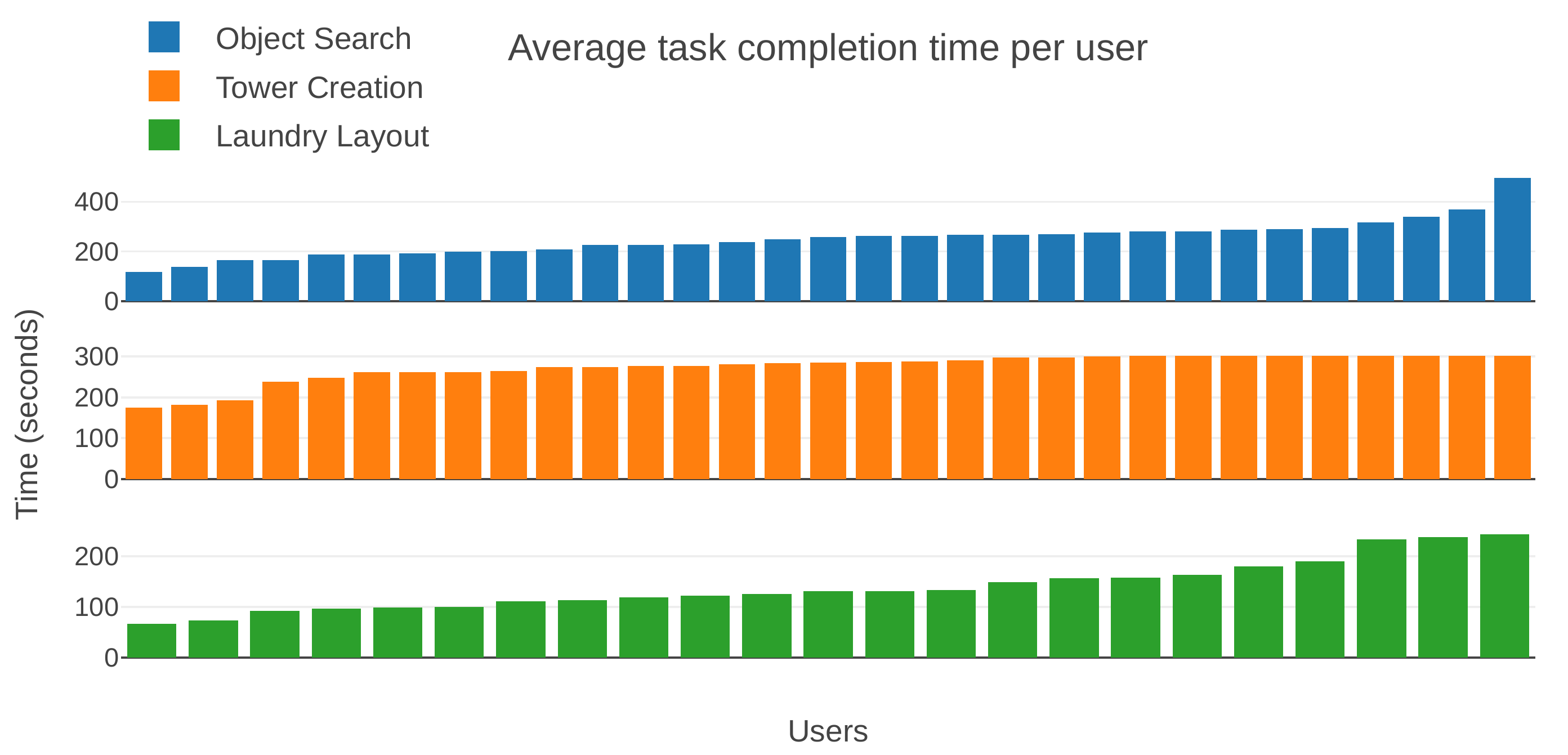}    
    \caption{\textbf{User skill comparison.} Average task completion time plotted for all users. The distribution indicates a wide range of skill exhibited across users on the Object Search and Laundry Layout tasks - more skillful users are able to complete these tasks in less time. By contrast, the uniform spread in times across the Tower Creation task indicates that all users were persistent in using all five minutes to create their tower.}
    \label{fig:time_per_user}
\end{figure}



The contributing users had a wide range of skill level, as seen in Fig.~\ref{fig:time_per_user}, but regardless of individual skill, we noticed strong trends of skill improvement with experience. We observed that users learned how to use the teleoperation control system very quickly, and once they became comfortable with it, they not only completed tasks faster, but more efficiently. Users demonstrated a higher degree of dexterity as they gained experience performing tasks.

Fig.~\ref{fig:experience_graphs} demonstrates quantitative results that corroborate what we observed. Task completion times steadily decreased as users gained experience doing the task. 
We also measured average user exertion, estimated by the square $L_2$ norms of phone translations during control, and found it largely invariant with respect to amount of experience, demonstrating that faster completion times were not due to users simply performing the same trajectory faster. 
By contrast, the average phone orientation change increased with experience, confirming that users learned to control the robot with more dexterous motion, enabling them to complete the task faster.



\subsection{Qualitative User Feedback}

Motivated by a previous study on robot teleoperation interfaces \cite{kent2017comparison}, we had each participant complete a NASA TLX form upon completion of the study \cite{hart1988development}.  This self-reported survey measured the participants' perception on mental demand, physical demand, temporal demand, performance, effort, and frustration on a 21-point scale. The total workload was computed as the sum of these averages, where higher scores represent a higher workload on users. From Table~\ref{tab:nasaTLS}, users found that the tower stacking task required the most workload across all the metrics.

\begin{table}[!t]
\centering
\vspace{1mm}
\caption{\textbf{NASA TLX Evaluation.} Self-reported user evaluation of the system.}
\resizebox{\linewidth}{!}{%
\begin{tabular}{lccccc}
\hline
\rowcolor[HTML]{CBCEFB}
\multicolumn{1}{l}{Measure} & Object Search & Tower Stacking  & Laundry Layout \\ \hline
\rowcolor[HTML]{EFEFEF}
\multicolumn{1}{l}{\textit{Mental demand}} & $12.2 \pm 3.7$ & $13.2 \pm 4.4$ & $9.3 \pm 5.3$  \\ 
\multicolumn{1}{l}{\textit{Physical demand}} & $11.1 \pm 4.0$ & $11.8 \pm 4.7$ & $9.6 \pm 4.6$   \\ 
\rowcolor[HTML]{EFEFEF}
\multicolumn{1}{l}{\textit{Temporal demand}} & $6.0 \pm 4.6$ & $12.2 \pm 6.2$ & $7.9 \pm 4.9$   \\ 
\multicolumn{1}{l}{\textit{Performance}} & $6.3 \pm 4.8$ & $12.4 \pm 5.1$ & $7.1 \pm 6.1$ \\ 
\rowcolor[HTML]{EFEFEF}
\multicolumn{1}{l}{\textit{Effort}} & $10.8 \pm 4.2$ & $14.2 \pm 3.9$ & $10.3 \pm 5.1$  \\ 
\multicolumn{1}{l}{\textit{Frustration}} & $7.5 \pm 5.0$ & $13.1 \pm 5.2$ & $7.3 \pm 5.4$ \\ \hline
\rowcolor[HTML]{EFEFEF}
\multicolumn{1}{l}{\textbf{Total Workload}}     & $53.9 \pm 11.2$ & $76.9 \pm 12.2$ & $51.5 \pm 12.8$    \\ 
\hline
\end{tabular}
}
\vspace{2mm}
\label{tab:nasaTLS}
\end{table}

\section{EXPERIMENTS: DATA ANALYSIS}

In this section, we investigate properties of our dataset that demonstrate the potential utility of the data for several applications such as multimodal density estimation, video prediction, reward function learning, policy learning and hierarchical task planning.

\subsection{Evaluating Object Search Task Complexity}

\begin{figure}[t]
    \centering
    \vspace{1mm}
    \includegraphics[width=\linewidth]{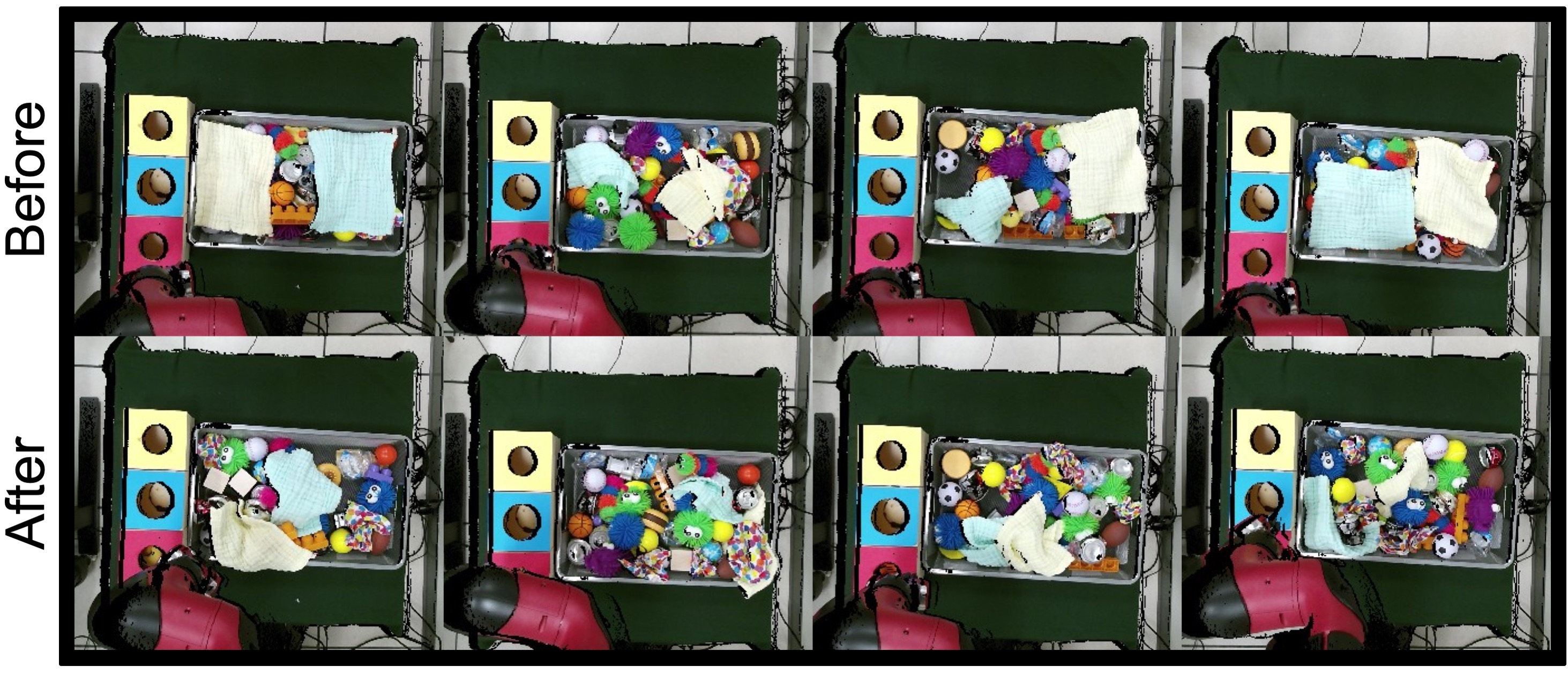}
    \caption{\textbf{Object Search Task Complexity.} The large difference between start and end frames on the Object Search demonstrates the significant amount of effort required to solve the task.}
    \label{fig:objsearch_effort}
\end{figure}

One of the key novelties of the dataset we present is the complexity of the reasoning necessary to plan a strategy for solving the tasks and the actual dexterity necessary to perform the finer details of manipulating the objects. We focus on the Object Search task since there is a simple qualitative measure to demonstrate the complexity of the task through the start and end configurations of the bin. Fig.~\ref{fig:objsearch_effort} shows the start and end configurations of the bin over several successful task demonstrations. The fact that these configurations vary greatly is evidence that the operator needed to drastically reorient the contents of the bin to find the objects of interest.

\subsection{Diversity of task demonstrations and solution approaches} 
\begin{figure*}[t!]
    \centering
    \includegraphics[width=0.98\linewidth]{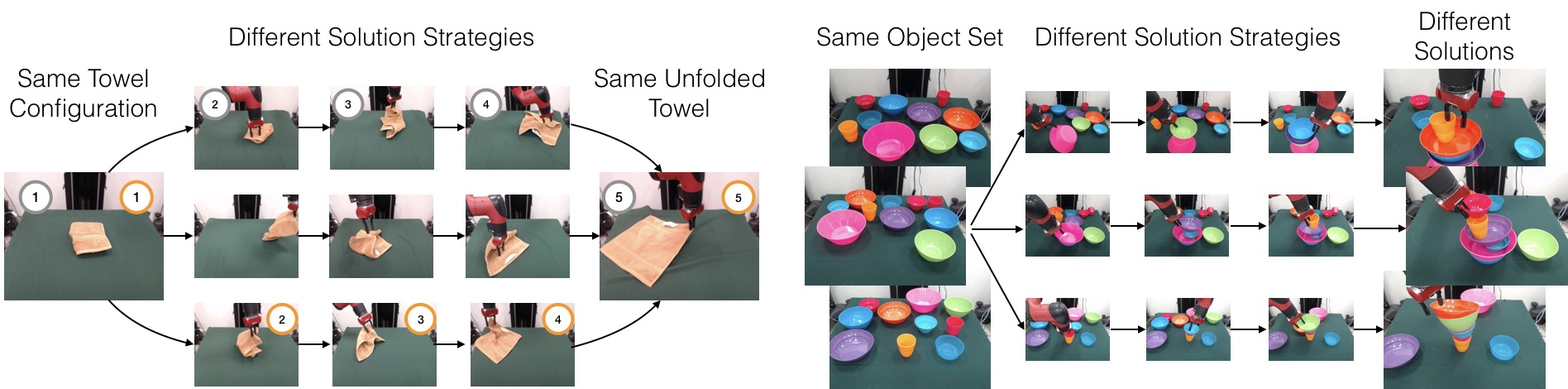}
    \vspace{5pt}
    \caption{\textbf{Diversity of Task Solutions.} Three different demonstrations provided on the Laundry Layout task (left) and the Tower Creation task (right) are shown above. The Laundry Layout demonstrations start and end with the same towel configuration across demonstrations but users exhibited different solution strategies to solve the task. The Tower Creation demonstrations start with the same set of objects in different configurations on the table but users chose to leverage the items in different ways, leading to three towers of roughly the same height, with different structural composition. This showcases the diversity of solution strategies that are present in our dataset.}
    \label{fig:multimodal}
\end{figure*}

One of the benefits of collecting data from a set of 54 users on tasks with an unstructured solution space is that every user has a unique approach to solve a given task instance. Fig.~\ref{fig:multimodal} presents three demonstrations on the Laundry Layout task that started with similar towel configurations and ended successfully with the towel flat on the table. The demonstration frames show that although the start and end configurations are similar, the approach used to flatten the towel are markedly different. For example, the user that provided the top demonstration chose to pick one side of the towel, place it down in a better configuration, then push the remaining flap of the towel open to complete the task. By contrast, the bottom demonstrator chose to grab a corner of the towel and manipulate it through lifting and dragging motions until the cloth was flat on the table. 

Fig.~\ref{fig:multimodal} also presents three demonstrations on the Tower Creation task that started with the same set of objects, but resulted in three completely different towers. While both towers were comparably high, the users demonstrated different ways to stack the cups and bowls in order to build the tower. 
The dataset that we collected contains many such instances of multimodality and creative problem solving that stem from the diversity of the humans that generated the data. 

\begin{figure}[t!]
    \centering
    \includegraphics[width=\linewidth]{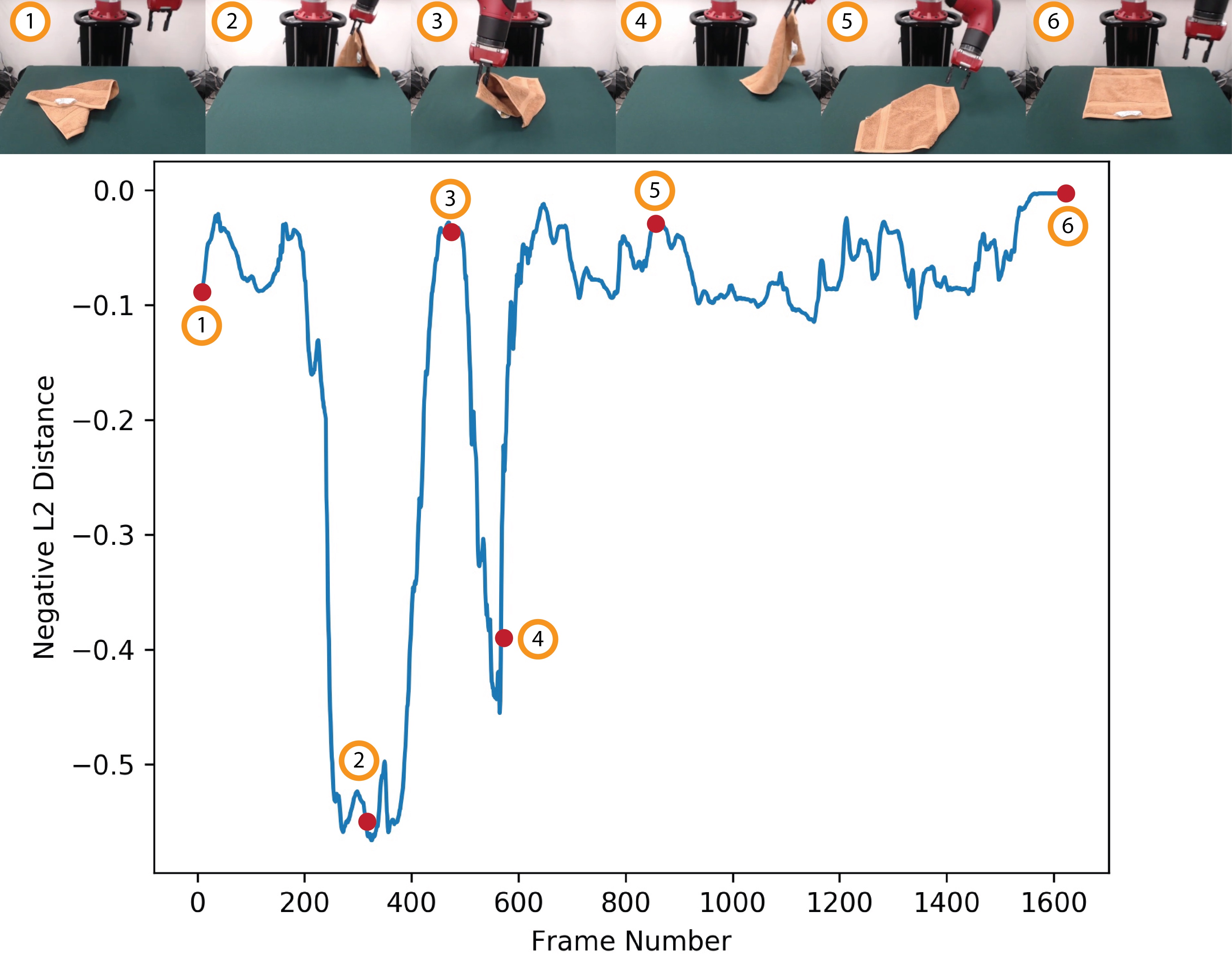}
    \caption{\textbf{Leveraging learned embedding spaces as a similarity metric for imitation.} We trained a custom variant of TCN~\cite{sermanet2018time} to learn an embedding space over RGB images on the Laundry Layout task. We plot the negative L2 embedding distance between a target frame with a flat towel (the last frame) and all other frames in a demonstrations. This distance provides a meaningful reward function for imitation as well as a useful metric for task progress.}
    \label{fig:reward}
    \vspace{3pt}
\end{figure}

\subsection{Inferring a Reward Signal from Demonstrations}

Consider the problem of learning a policy to imitate a specific video demonstration. Prior work \cite{sermanet2018time, aytar2018playing} has approached this problem by learning an embedding space over visual observations and then crafting a reward function to imitate a reference trajectory based on distances in the embedding space. This reward function can then be used with reinforcement learning to learn a policy that imitates the trajectory. Taking inspiration from this approach, we trained a modified version of Time Contrastive Networks (TCN)~\cite{sermanet2018time} on Laundry Layout demonstrations and investigate some interesting properties of the embedding space. 

To address the large and diverse amount of data that was collected, we made two important modifications to the TCN algorithm. The original algorithm used a triplet loss to encourage neighboring video frames to be close in the embedding space; however, we found that applying the original TCN algorithm to our dataset resulted in embeddings with distances that were not meaningful for frames with larger time separation in a demonstration. Learning an embedding space that can tolerate frames with large temporal separation is critical for our dataset, since our tasks are multi-stage and our demonstrations are several minutes long.

In order to learn both high and low frequency temporal similarity, we split each demonstration into \textit{chunks} of uniform size and use two separate triplet losses - an intra-chunk loss that pushes neighboring frames from within the same chunk of time together in the embedding space and an inter-chunk loss that encourages frames from nearby chunks of time to be close in the embedding space. We also added an auxiliary loss to encourage terminal demonstration frames to be close in the embedding space.

In Fig.~\ref{fig:reward}, we consider the frame embeddings along a single Laundry Layout demonstration. We plot the negative L2 distance of the frame embeddings with respect to the embedding of a target frame near the end of the video, where the target frame depicts a successful task completion with the towel lying flat on the table. 
The figure demonstrates that distances in this embedding space with a suitable target frame yield a reasonable reward function that could be used to imitate task demonstrations purely from visual observations. 

Furthermore, embedding distances capture task semantics to a certain degree and could even be used to measure task progress. For example, in frames 3 and 5, the towel is nearly flat on the table, and the embedding distance to frame 6 is correspondingly small. By contrast, in frames 2 and 4, the robot is holding the towel a significant distance away from the table, and the distance to frame 6 is correspondingly large.    

\subsection{Behavioral Cloning}

We also trained policies using Behavioral Cloning on the Laundry Layout task by learning a mapping from RGB images to robot joint positions. Our attempts to learn from the entire dataset were ultimately unsuccessful due to the diverse nature of the demonstrations, but we were able to achieve some success by restricting the training data to demonstration segments where the arm moves to a corner of the towel, and lifts the towel up. Addressing the diversity of the dataset for policy learning is left for future work.


\section{CONCLUSION}

We introduced three challenging manipulation tasks: Object Search, Tower Creation, and Laundry Layout. Solving each of these tasks requires both higher-level reasoning to specify \textit{what} to accomplish and dexterous manipulation that answers \textit{how} to accomplish the necessary physical interactions. Each task instance also admits many diverse solutions, making these tasks amenable to crowdsourcing.

We presented the largest known crowdsourced teleoperated robot manipulation dataset consisting of over 111 hours of data across 54 users. The dataset was collected in 1 week on 3 Sawyer robot arms using the RoboTurk platform. We made important extensions to the RoboTurk platform to enable data collection on physical robots, including accounting for additional delays in the remote teleoperation loop due to physical robot actuation, ensuring operational safety of the robots when being controlled by novices, and managing large-scale data collection across multiple sensor streams. We analyzed how the system allowed our participants to adapt quickly to the phone control interface in order to collect diverse, successful demonstrations on the three tasks. We also presented a set of qualitative and quantitative results that showcase the diversity and utility of the dataset. Our dataset can be useful for several applications such as multimodal density estimation, video prediction, reward function learning, policy learning and hierarchical task planning.


Future work will be focus on two directions - improving the platform and utilizing the dataset for policy learning. Platform improvements include: (1) alleviating the need for manual task resets by collecting data on reversible tasks (a forward task and a reset task) so that remote operators can reset the workspace and (2) developing a more structured scheduling system for operators that ensures fair waiting times for those in the queue. While our initial results suggest that the dataset can potentially be used for both one-shot imitation learning and direct imitation, further considerations and innovations will be necessary to handle the inherent diversity and multimodality of the solutions demonstrated. 

\vspace{1mm}
{\small
\noindent
\textbf{Acknowledgement.}
Ajay Mandlekar is supported by the Department of Defense (DoD) through the NDSEG program.
We acknowledge the support of ONR-MURI (1175361-6-TDFEK), Toyota Research Institute (S-2015-09-Garg), Weichai America Corp. (1208338-1-GWMZY), CloudMinds Technology Inc. (1203462-1-GWMXN), Nvidia (1200225-1-GWMVU), Toyota (1186781-31-UBLFA), Panasonic, and Google Cloud. 
We would like to thank the users that contributed to the dataset, as well as the Stanford People, AI \& Robots (PAIR) group, and the anonymous reviewers for their constructive feedback.}

\newpage
\renewcommand*{\bibfont}{\footnotesize}
\begin{flushright}
\printbibliography 
\end{flushright}

\newpage
\section{Appendix}

\subsection{Additional Task Details}

\begin{figure}[h!]
    \centering
    \includegraphics[width=0.7\linewidth]{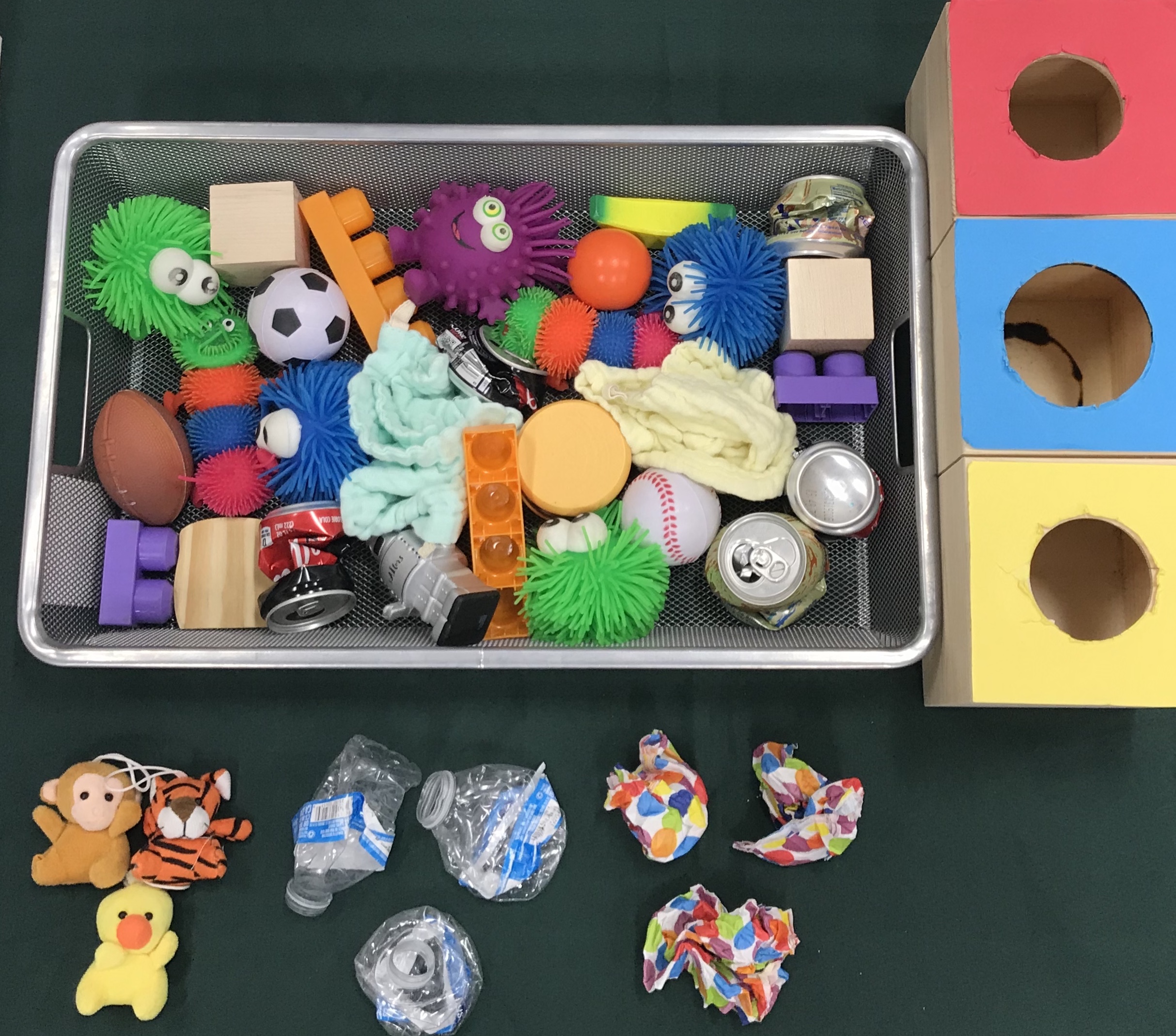}
    \caption{Items used for the Object Search Task. Target objects are shown outside of the bin.}
    \label{fig:app_object_search}
\end{figure}

\noindent \textbf{Object Search.} Fig.~\ref{fig:app_object_search} shows the target object categories and items and the other objects present in the cluttered workspace. At the beginning of each task instance, the operator was provided with one of the three target object categories (\textit{plush animals}, \textit{plastic water bottles}, and \textit{paper napkins}) at random, and the objective was to place all three items of that category into the appropriate hole. The \textit{plush animals} had to be placed in the red hole, \textit{plastic water bottles} had to be placed in the blue hole. and the \textit{paper napkins} had to be placed in the yellow hole. Note that all objects were present in the clutter at the start of every task instance, including the target objects of all categories, and the contents of the clutter were shuffled between demonstrations to ensure diverse task instances.

\begin{figure}[h!]
    \centering
    \includegraphics[width=0.7\linewidth]{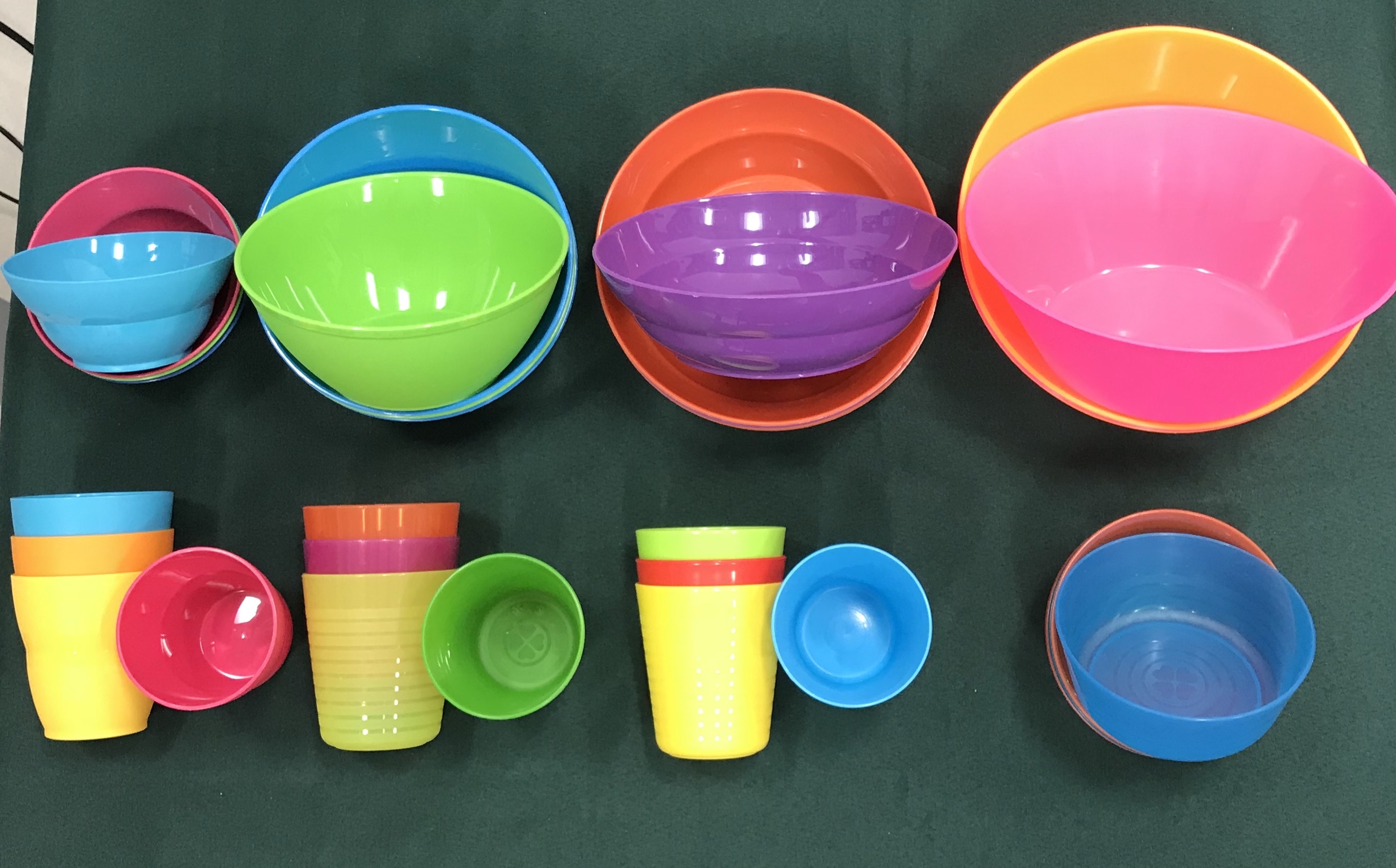}
    \caption{Item categories used for the Tower Creation task. Eight different categories of bowls and cups were used, with four items in each category.}
    \label{fig:app_tower_creation}
\end{figure}

\noindent \textbf{Tower Creation.} Fig.~\ref{fig:app_tower_creation} shows the complete set of 28 bowls in 7 varieties and 12 cups in 3 varieties used to generate task instances. Every task instance consisted of a random assortment of 10 items sampled from this set, laid out in a random configuration on the table.

\begin{figure}
    \centering
    \includegraphics[width=0.7\linewidth]{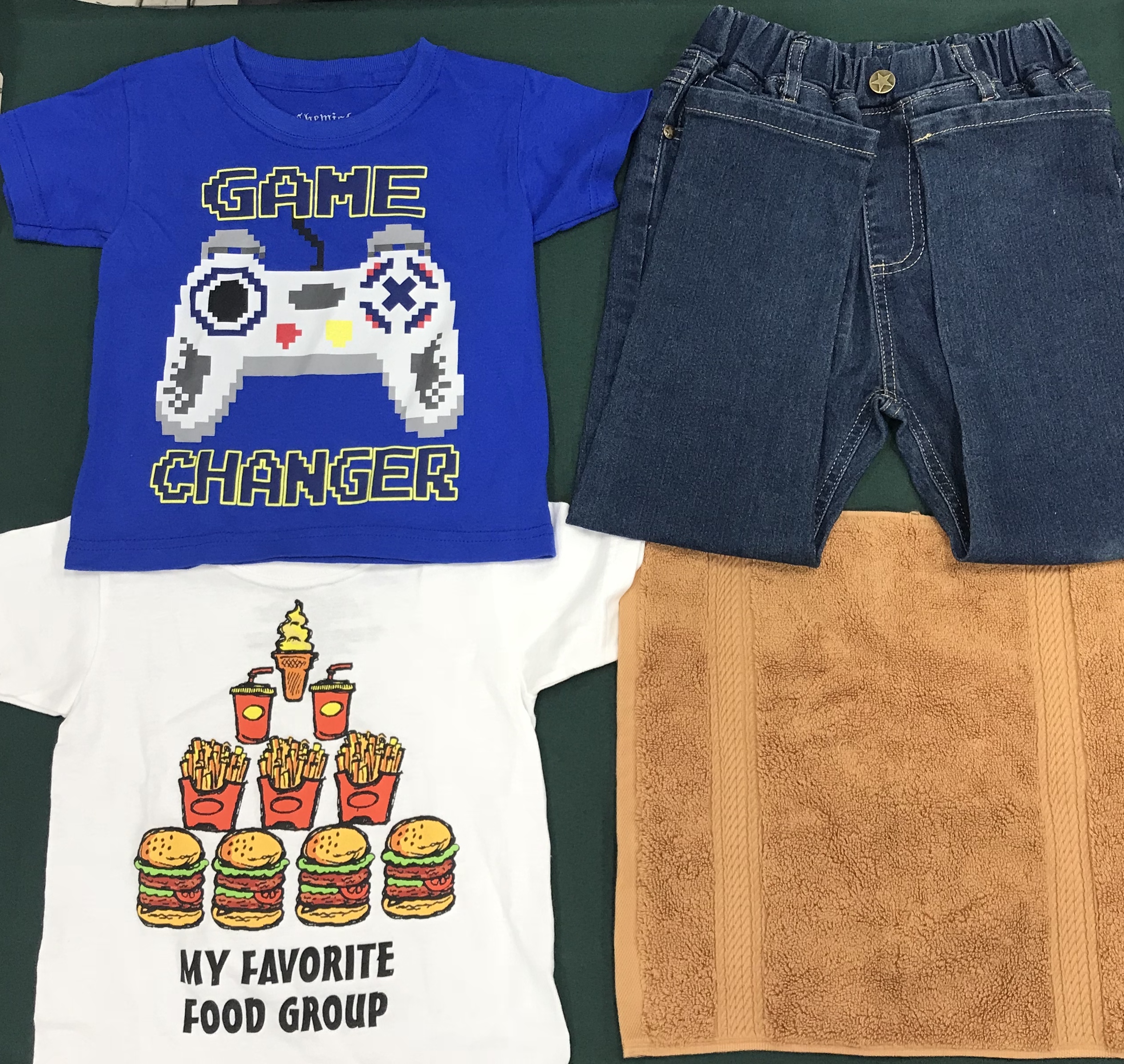}
    \caption{Items used for the Laundry Layout task. The items consist of two T-shirts (left), a pair of jeans (top right) and a hand towel (bottom right).}
    \label{fig:app_laundry_layout}
\end{figure}

\noindent \textbf{Laundry Layout.} Fig.~\ref{fig:app_laundry_layout} shows the hand towels and clothing used between task instances. New operators were always given task instances with hand towels, while more experienced operators were given more challenging instances with jeans and t-shirts.

\subsection{Reward Learning Experiment Details}
In this section, we describe details on how video demonstrations were used to learn an embedding space suitable for crafting reward functions for imitation learning. In order to effectively learn reward functions from large amounts of demonstrations consisting of long sequences, we use a hierarchical approach by extending Time Contrastive Networks (TCN)~\cite{sermanet2018time}. We split each demonstration into \textit{chunks} of approximately $1$ second in length ($24$ frames) as in the original TCN method. We then train a model using the triplet loss over time where we sample anchors, positive, and negative examples from within each chunk. Specifically we define
\[
    L_{\text{high-freq}} = L_{triplet}(a, p, n)
\]
where $a$ is an anchor frame, $p$ is a positive example, and $n$ is a negative example sampled within a given chunk with the positive radius of $6$ and a negative radius of $12$. We also introduce a second loss function that encourages chunks that are close in time to also be close in embedding space with another triplet loss which we define as 
\[
    L_{\text{low-freq}} = L_{triplet}(a, p, n)
\]
where we sample $a$ randomly from a chunk and $p$ and $n$ at random from chunks in radii $3$ and $6$ respectively. This allows the model to learn a relationship between frames at a coarser temporal resolution. We then optimize over the loss
\[
    L_{TCN} = \lambda_1 L_{high_freq} + \lambda_2 L_{low_freq}
\]
where we chose $\lambda_1 = \lambda_2 = 1$. In practice, we use semi-hard negative mining to improve the optimization of the triplet loss. Our embedding network consists of a typical ResNet-18 with output dimension of $1000$ and add an additional two fully connected layers with sizes $256$ and $32$. The loss functions are optimized using Adam with a learning rate of $1e-3$.

\subsection{Data Diversity}
\begin{figure*}[h!]
    \centering
    \includegraphics[width=0.7\linewidth]{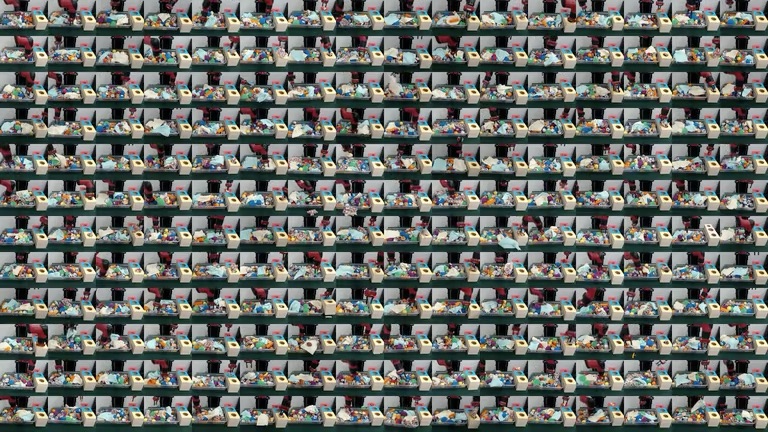}
    \caption{A random assortment of frames sampled from the dataset on the Object Search task.}
    \label{fig:mosaic1}
\end{figure*}

\begin{figure*}[h!]
    \centering
    \includegraphics[width=0.7\linewidth]{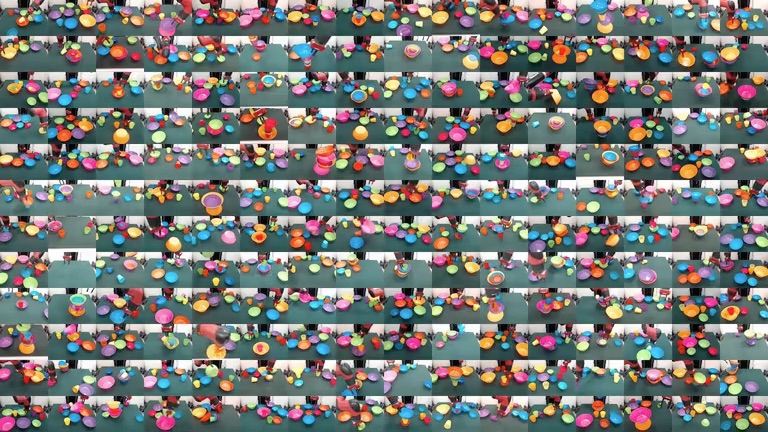}
    \caption{A random assortment of frames sampled from the dataset on the Tower Creation task.}
    \label{fig:mosaic2}
\end{figure*}

\begin{figure*}[h!]
    \centering
    \includegraphics[width=0.7\linewidth]{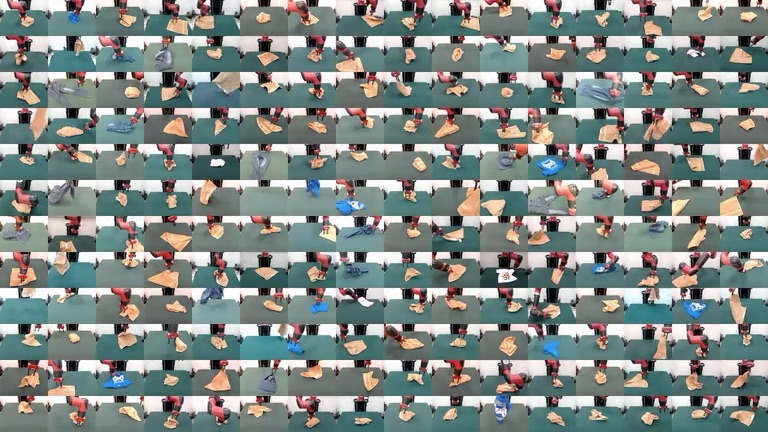}
    \caption{A random assortment of frames sampled from the dataset on the Laundry Layout task.}
    \label{fig:mosaic3}
\end{figure*}

We present an assortment of frames sampled from our dataset in Fig.~\ref{fig:mosaic1}, Fig.~\ref{fig:mosaic2}, and Fig.~\ref{fig:mosaic3}.

\end{document}